\documentclass[acmtog,nonacm]{acmart}

\usepackage{epsfig}
\usepackage{graphicx}
\usepackage{amsmath}

\usepackage{booktabs}
\usepackage{multirow}
\usepackage{bigdelim}

\usepackage{geometrycollective}

\usepackage{arydshln}

\newtheorem{theorem}{Theorem}
\newtheorem{lemma}[theorem]{Lemma}

\newcommand{\nodata}{\multicolumn{1}{c}{---}}

\usepackage{array}
\newcolumntype{L}[1]{>{\raggedright\let\newline\\\arraybackslash\hspace{0pt}}m{#1}}
\newcolumntype{C}[1]{>{\centering\let\newline\\\arraybackslash\hspace{0pt}}m{#1}}
\newcolumntype{R}[1]{>{\raggedleft\let\newline\\\arraybackslash\hspace{0pt}}m{#1}}

\newcommand{\bftab}{\fontseries{b}\selectfont}

\setcopyright{acmcopyright}
\copyrightyear{2021}
\acmYear{2021}
\acmDOI{10.1145/9999999.9999999}

\acmJournal{TOG}
\acmVolume{99}
\acmNumber{99}
\acmArticle{99}
\acmMonth{99}

\begin{document}

\title{DiffusionNet: Discretization Agnostic Learning on Surfaces}

\author{Nicholas Sharp}
\affiliation{%
  \institution{Carnegie Mellon University, University of Toronto}
}

\author{Souhaib Attaiki}
\affiliation{%
  \institution{LIX, \'{E}cole Polytechnique}
}

\author{Keenan Crane}
\affiliation{%
  \institution{Carnegie Mellon University}
}

\author{Maks Ovsjanikov}
\affiliation{%
  \institution{LIX, \'{E}cole Polytechnique}
}

\renewcommand{\shortauthors}{Sharp et al.}

\begin{abstract}
  \vspace*{2em}
  We introduce a new general-purpose approach to deep learning on 3D surfaces, based on the insight that a simple diffusion layer is
  highly effective for spatial communication.  The resulting networks are automatically robust to changes in
  resolution and sampling of a surface---a basic property which is crucial for practical applications.  Our networks
  can be discretized on various geometric representations such as triangle meshes or point clouds, and can even be
  trained on one representation then applied to another.  We optimize the spatial support of diffusion as a continuous
  network parameter ranging from purely local to totally global, removing the burden of manually choosing neighborhood
  sizes.  The only other ingredients in the method are a multi-layer perceptron applied independently at each point, and
  spatial gradient features to support directional filters.  The resulting networks are simple, robust, and efficient.
  Here, we focus primarily on triangle mesh surfaces, and demonstrate state-of-the-art results for a variety of tasks
  including surface classification, segmentation, and non-rigid correspondence.
  \vspace{2em}
\end{abstract}

\ccsdesc[500]{Computing methodologies~Shape analysis}

\keywords{geometric deep learning, geometry processing, discrete differential geometry, partial differential equations}

\maketitle

\vspace*{3em}
\section{Introduction}
\label{sec:intro}

Recently there has been significant interest in learning techniques for non-uniform geometric data, inspired by the tremendous success of convolutional neural networks (CNNs) in computer vision.
A particularly challenging setting is extending the power of CNNs to learning directly on curved surfaces
\cite{masci2015geodesic,bronstein2017geometric,poulenard2018multi,hanocka2019meshcnn}.
Unlike volumetric \cite{maturana2015voxnet} or point-based \cite{qi2017pointnet} approaches, surface-based methods exploit the connectivity of the surface representation to improve performance, and furthermore can be robust in the presence of non-rigid deformations, making them a strong solution for many tasks such as deformable shape matching \cite{masci2015geodesic,boscaini2016learning}.

However, although the field has largely been focused on the benchmark accuracy of networks for such problems, at least two other major roadblocks remain for achieving the full potential of learning on surfaces.
First, whereas real-world geometric data comes from a variety of sources, existing networks are strongly tied to a particular representation (\eg{}, triangulations \emph{or} point clouds) or even discretization resolution.  Hence, training cannot benefit from all available data.
One popular strategy is to simply convert all data to a common representation (\eg{}, via point sampling), but this approach has well-known drawbacks (sampling a high-quality, detailed surface can alias thin features, lose informative details, \etc{}).
Second, many existing mesh-based architectures do not scale well to high-resolution surface data.
Though coarse inputs are sufficient for, \eg{}, classification tasks, they preclude potential future applications such as high-fidelity geometric analysis and synthesis.

\begin{figure}
\begin{center}
\end{center}
    \includegraphics{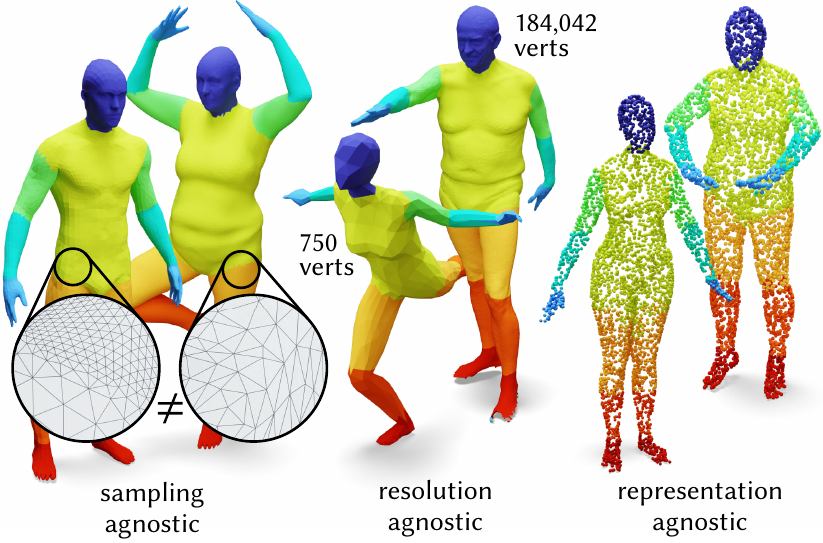}
    \caption{
      Surface learning methods must generalize to shapes represented differently from the training set to be useful in practice, yet many existing approaches depend strongly on mesh connectivity.
      Here, our DiffusionNet trained for human segmentation with limited variability seen during training automatically generalizes to widely varying mesh samplings (\figloc{left}), scales gracefully to resolutions ranging from a simplified model to a large raw scan (\figloc{middle}), and can even be evaluated directly on point clouds (\figloc{right}).
    \label{fig:generalization_teaser}}
\end{figure}

A key technical difficulty in surface-based learning is defining appropriate notions of convolution and pooling---two main building blocks in traditional CNNs. Unfortunately, unlike the Euclidean case, there is no universal canonical notion of convolution on surfaces.
Existing approaches have tried to address this challenge  through a variety of solutions such as mapping to a canonical domain \cite{sinha2016deep,maron2017convolutional}, exploiting local parametrizations \cite{masci2015geodesic,boscaini2016learning,wiersma2020cnns} or applying convolution on the edges of the mesh \cite{hanocka2019meshcnn}.
However, the use of more advanced and delicate geometric operations, such as computing geodesics or parallel transport, has a significant impact on both the robustness and scalability of the resulting methods.
Perhaps even more importantly, existing surface-based approaches are often \emph{too} sensitive to the underlying mesh structure, and thus unable to generalize to significantly different sampling and triangulations between training and test sets. As a result, despite significant recent progress in geometric deep learning \cite{cao2020comprehensive,greengard2021}, current methods typically struggle to cope with the variability, complexity and scale encountered in the real-world surface and mesh-based settings.

In this work, we propose a method that exploits the surface representation, but is both scalable and robust in the presence of significant sampling changes (see Figure \ref{fig:generalization_teaser}). Our main observation is that expensive and potentially brittle operations used in previous works \cite{masci2015geodesic,poulenard2018multi,wiersma2020cnns} can be replaced with two basic geometric operations: a learned diffusion layer for information propagation and a spatial gradient for capturing anisotropy. Discretizing these operations with the principled techniques of discrete differential geometry~\cite{meyer2003discrete,crane2013digital} then \emph{automatically} endows the resulting networks with both robustness and scalability, while maintaining the simplicity of the learning framework.

Remarkably, we show that combining these basic geometric operations yields neural networks that are not only robust and scalable, but also achieve state-of-the-art results in a wide variety of applications, including deformable surface segmentation, classification, as well as unsupervised and supervised non-rigid shape matching. Perhaps even more fundamentally, our DiffusionNet offers a unified perspective across representations of surface geometry---in principle it can be applied to any geometric representation where one has a Laplacian and gradient operator.
In this paper, for example, we show how the same architecture achieves accurate results for both meshes and point clouds, and even allows training on one and evaluating on the other.

\paragraph{Contributions.}
The main contributions of this work are:
\begin{itemize}
  \setlength\itemsep{0.2em}
   \item We show that a simple learned diffusion operation is sufficient to share spatial data in surface learning.
   \item We introduce spatial gradient features for learning local directional filters.
   \item Inspired by these insights, we present DiffusionNet, an architecture for learning on surfaces which has many advantages including robustness to discretization, and achieves state-of-the-art results on several benchmarks.
\end{itemize}

\section{Related Work}
Applying deep learning techniques to 3D shapes is a rich and extensive area of research. Below we review the approaches most closely related to ours, and refer the interested readers to recent surveys, including \cite{xu2016data,bronstein2017geometric,cao2020comprehensive}.

\vspace{-1mm}
\paragraph{View-based and volumetric methods.}
Most early geometric deep learning-based methods directly leveraged tools developed for 2D images, and thus mapped 3D shapes onto the plane either using multi-view renderings \cite{su2015multi,wei2016dense,kalo17} or more global, often parametrization-based techniques such as panoramas
\cite{shi2015deeppano,sfikas2017exploiting}, geometry images \cite{sinha2016deep}, or metric-preserving mappings \cite{ezuz2017gwcnn}, among many others.

Another direct approach to applying convolution to 3D shapes relies on volumetric voxel grid representations, which has led to a variety of methods, including \cite{wu20153d,maturana2015voxnet} and their efficient extensions
\cite{wang2017cnn,klokov2017escape}. Such techniques can, however, be computationally expensive and difficult to apply to detailed deformable shapes.

\subsection{Learning on Surfaces}
Methods that learn on 3D surfaces directly typically fall into two major categories, based either on point cloud or triangle mesh representations.

\vspace{-1mm}
\paragraph{Point-based methods.}
A successful set of methods for learning on 3D shapes represented as point clouds was pioneered by the PointNet  \cite{qi2017pointnet} and
PointNet++ \cite{qi2017pointnet++} architectures, which have been extended in many recent works, including PointCNN  \cite{li2018pointcnn}, DGCNN \cite{wang2019dgcnn}, PCNN \cite{atzmon2018pcnn} and KPConv \cite{thomas2019kpconv} to name a few (see also \cite{guo2020deep} for a recent survey). Moreover, recent efforts have also been made to incorporate invariance and equivariance of the networks with respect to various geometric transformations, \eg{}, \cite{deng2018ppf, poulenard2019effective,zhang2019rotation,zhao2020quaternion,li2021rotation,hansen2018multi}. The major advantages of point-based methods are their simplicity, flexibility, applicability in a wide range of settings and robustness in the presence of noise and outliers.
However, first, their overall accuracy can often be lower than that of methods that explicitly use surface (\eg{}, mesh) connectivity when it is available.
Second, though effective on static mechanical objects and scenes, point-based methods may not be well-suited for \emph{deformable} (non-rigid) shape analysis, requiring extremely large training sets and significant data augmentation to achieve good results, \eg{}, for non-rigid shape matching applications \cite{groueix2018b,Donati2020DeepGF}.
Globally supported point-based networks were recently considered in~\cite{peng2020efficient}; our method naturally allows global support via learned diffusion.

\begin{figure}[b]
\begin{center}
\end{center}
    \includegraphics{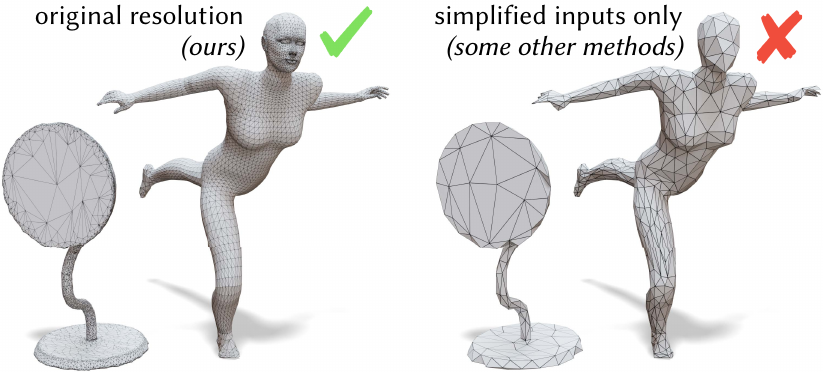}
    \caption{
      Many recent mesh-based learning methods are applied only to dramatically simplified inputs (\secref{performance}), while our method easily processes full-resolution models, preserving detail and facilitating adoption.
      \label{fig:resolution}}
\end{figure}

\vspace{-1mm}
\paragraph{Surface and graph-based methods.}
To address the limitations of point-based approaches, several methods have been proposed that operate directly on mesh surfaces and thus can learn filters that are intrinsic and robust to complex non-rigid deformations. The earliest pioneering approaches in this direction generalize convolutions~\cite{masci2015geodesic,boscaini2016learning,monti2017geometric,fey2018splinecnn}, typically using local surface parameterization via the logarithmic map.
Unfortunately local parameterizations are only defined up to rotation in the tangent plane, leading to several methods which address this issue through design of \emph{equivariant} surface networks~\cite{poulenard2018multi,he2020curvanet,wiersma2020cnns,yang2021continuous, de2020gauge,mitchel2021field}.
Likewise, operating on vector-valued data in a local tangent space can expand the expressivity of the filter space~\cite{wiersma2020cnns,mitchel2021field,beani2021directional}.
Our method leverages learned gradient features (\secref{spatial_gradients}), which geometrically require only a local spatial gradient operation, and sidestep the challenge of equivariant filters by using only inner products, which are naturally invariant.
These gradient features are built on local differential operators, which have also been exploited in other recent methods (\eg{}, ~\cite{jiang2018spherical,eliasof2020diffgcn}).

\begin{figure}[b]
\begin{center}
\end{center}
    \includegraphics{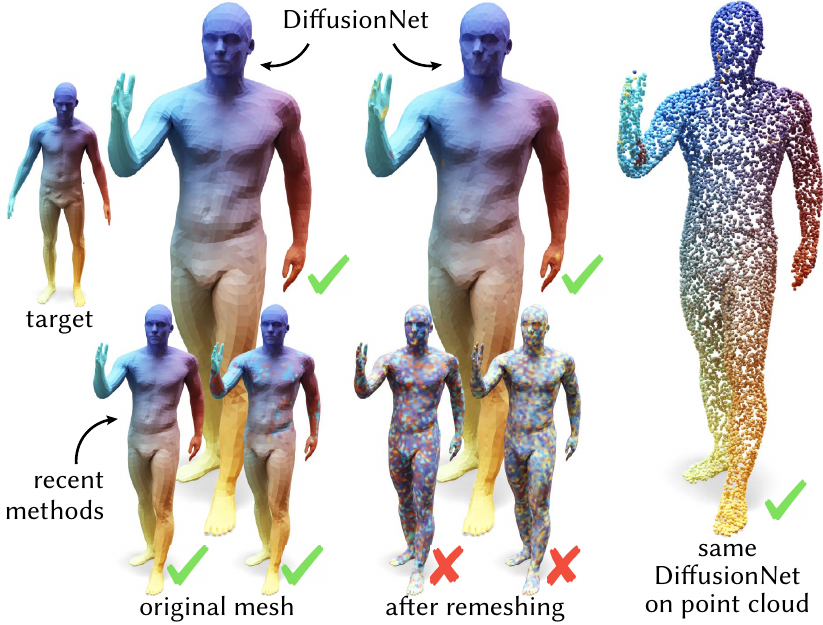}
    \caption{
      Although past methods have achieved high-accuracy benchmark results for learning on meshes~\cite{li2020shape, fey2018splinecnn}, they are prone to over-fitting to the mesh connectivity, rather than learning the underlying shape structure (\secref{sampling_invariance}).
      In contrast, DiffusionNet learns an accurate representation-agnostic solution, which even supports training on meshes and evaluating on a point cloud (\emph{last column}).
    \label{fig:remesh_fail}}
\end{figure}

Surface mesh structure has also been used in variety of graph-like approaches which specifically leverage mesh connectivity \cite{hanocka2019meshcnn,verma2018feastnet,lim2018simple,gong2019spiralnetpp,feng2019meshnet, milano2020primaldual, hajij2020cell, bodnar2021weisfeiler}, the structure of discrete operators~\cite{smirnov2021hodgenet}, or even random walks along edges~\cite{lahav2020meshwalker}, among others. While accurate, these methods can be costly on densely sampled shapes and often are not robust to significant changes in the mesh structure (\figref{remesh_fail}).

\subsection{Spectral Methods}
Our use of diffusion is also loosely related to techniques that operate in the spectral domain and often exploit the link between convolution and operations in a derived (\eg{}, Fourier or Laplace-Beltrami) basis, including \cite{bruna2014,levie2018cayleynets,sun2020zernet}. Such methods have a long history in graph-based learning and are well-rooted in data analysis more broadly, including Laplacian eigenmaps \cite{belkin2003laplacian}, spectral clustering \cite{vallet2008spectral} and diffusion maps \cite{coifman2005geometric}.
In geometry processing, spectral methods have been used for a range of tasks including multi-resolution representation \cite{levy2006laplace}, segmentation \cite{rustamov2007} and matching \cite{ovsjanikov2012functional}, among others \cite{vallet2008spectral,zhang2010spectral}.

Unfortunately, Laplacian eigenfunctions depend on each shape and thus coefficients or learned filters from one shape are not trivially transferable to another.
\citet{levie2019transferability} argue for transfer between discretizations of the same shape, but 3D geometric learning typically demands transfer between different shapes.
Functional maps \cite{ovsjanikov2012functional} can be used to ``translate'' coefficients between shapes, and have been used, \eg{}, in \cite{yi2017syncspeccnn} with spectral filter learning. Deep functional maps \cite{litany2017deep} propose to learn features in the primal domain, which are then projected onto the Laplace-Beltrami basis for functional map estimation. However, the features are still learned either with MLPs starting from pre-computed descriptors \cite{litany2017deep,roufosse2019unsupervised,ginzburg2020cyclic,halimi2019unsupervised} or using point-based architectures \cite{Donati2020DeepGF}.

Instead, we propose an approach that learns the parameters of a diffusion process which is directly transferable across shapes and, as we show below, can be used effectively in applications like non-rigid shape matching.
We also stress that DiffusionNet \emph{is not} spectral in nature and only uses spectral operations as an acceleration technique for evaluating diffusion efficiently.

We also note that our use of the Laplacian in defining the diffusion operator is related to methods based on polynomials of the Laplacian~\cite{kostrikov2018surface,defferrard2016convolutional}, CayleyNets  \cite{levie2018cayleynets} and their recent application in shape matching using ACSCNNs \cite{li2020shape}. However, we demonstrate that complex polynomial filters can be replaced with simple learned diffusion, and moreover that gradient features can inject orientation information into the network, improving performance and robustness.

Similarly to our approach, diffusion for smooth communication has been explored on graphs~\cite{klicpera2019diffusion,xu2020graph}, images \cite{liu2016learning}, and point clouds~\cite{hansen2018multi}.
In contrast, our method directly learns a diffusion time per-feature (which significantly improves performance, \tabref{ablation}), incorporates a learned gradient operation, and is applied directly to mesh surfaces.

\vspace{-2mm}
\paragraph{Pooling}
In surface learning it is nontrivial to define pooling, especially on meshes where it often amounts to mesh simplification \cite{hanocka2019meshcnn}. Various recent operations have been proposed for point cloud \cite{Lin_2020_CVPR,Hu2020}, mesh \cite{milano2020primaldual,zhou2020fully} or even graph pooling \cite{ma2020path,li2020graph}. A key advantage of our approach is that it automatically supports global spatial support without any downsampling operation, simplifying implementation and improving learning.

\section{Method}
\label{sec:method}

Our method consists of three main building blocks: multi-layer perceptrons (MLPs) applied at each point to model pointwise scalar functions of feature channels, a learned diffusion operation for propagating information across the domain, and local spatial gradient features to expand the network's filter space beyond radiallly-symmetric filters.
In this section, we describe these main numerical components, and then we assemble them into an effective architecture in \secref{architecture}.
Our method is defined in a representation-agnostic manner; applying it to meshes or point clouds simply amounts to assembling the appropriate Laplacian and gradient matrices as we discuss below.

\subsection{Pointwise Perceptrons}

On a mesh or point cloud with $V$ vertices, we consider a collection of $D$ scalar features defined at each vertex.
Our first basic building block is a pointwise function $f : \mathbb{R}^D \to \mathbb{R}^D$, which is applied independently at every vertex to transform the features.
We represent these pointwise functions as a standard multilayer perceptron (MLP) with shared weights across all vertices. Although these MLPs can fit arbitrary functions at each point, they do not capture the spatial structure of the surface, or allow any communication between vertices, so a richer structure is needed.

Past approaches for communication have ranged from global reductions to explicit geodesic convolutions---instead, we will demonstrate that a simple learned diffusion layer effectively propagates information, without the need for potentially costly or error-prone computations.

\subsection{Learned Diffusion}
\label{sec:learned_diffusion}

In the continuous setting, diffusion of a scalar field $u$ on a domain is modeled by the \emph{heat equation}
\begin{equation}
  \label{eq:diffusion}
  \tfrac{d}{dt} u_t = \Delta u_t,
\end{equation}
where \(\Delta\) is the Laplacian (or more formally: the \emph{Laplace-Beltrami operator}).
The action of diffusion can be represented via the heat operator $H_t$, which is applied to some initial distribution $u_0$ and produces the diffused distribution $u_t$; this action can be defined as $H_t(u_0) = \exp(t \Delta) u_0$, where $\exp$ is the operator exponential.
Over time, diffusion is an increasingly-global smoothing process: for \(t=0\), $H_t$ is the identity map, and as \(t \to \infty\) it approaches the average over the domain.

We propose to use the heat equation to spatially propagate features for learning on surfaces; its principled foundations ensure that results are largely invariant to the way the surface is sampled or meshed.
To discretize diffusion, one replaces \(\Delta\) with the weak Laplace matrix $L$ and mass matrix $M$. Here, $L$ is a positive semi-definite sparse matrix $L \in \mathbb{R}^{V \times V}$ with the opposite sign convention such that $M^{-1} L\!\approx\!-\Delta$.
The number of entries in $L$ and $M$ are generally $O(V)$, scaling effectively to large inputs (\tabref{runtime}).
On triangle meshes, we will use the \emph{cotan-Laplace} matrix, which is ubiquitous in geometry processing applications~\cite{MacNeal:1949:SPD,pinkall1993computing,crane2013digital}; for point clouds we will use the related Laplacian from~\cite{sharp2020laplacian}.
This matrix has also been defined for voxel grids~\cite{caissard2019laplace}, polygon meshes~\cite{bunge2020polygon}, tetrahedral meshes~\cite{alexa2020properties}, \etc{}\,
The weak Laplace matrix is accompanied by a mass matrix $M$, such that the rate of diffusion is given by $-M^{-1} L u$.
Here $M$ will be a ``lumped'' diagonal matrix of areas associated with each vertex.

\begin{figure}
\begin{center}
\end{center}
    \includegraphics{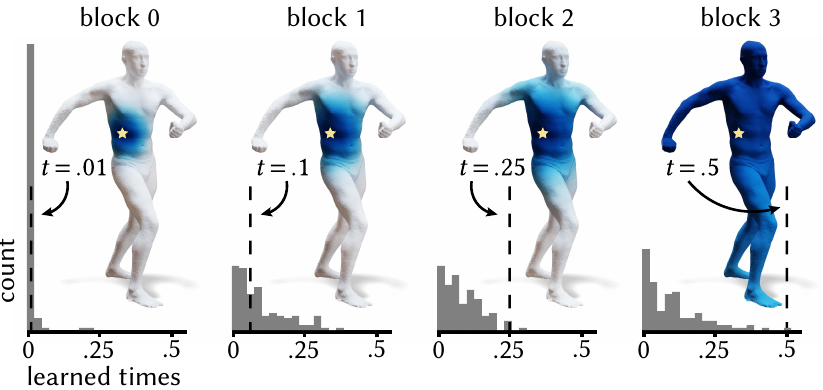}
    \caption{
    We propose to learn a diffusion time for each feature channel, automatically tuning spatial support during training.
    The histograms show the learned times at each block in a DiffusionNet trained for segmentation; the times marked by the dashed lines are visualized by diffusing a point source from the starred point.
    The first block uses mainly local diffusion, while a channel in the last block finds nearly global support.
    \label{fig:diffusion_viz}}
\end{figure}

We define a learned diffusion layer $h_t : \mathbb{R}^{V} \to \mathbb{R}^{V}$, which diffuses a feature channel $u$ for learned time $t \in \mathbb{R}_{\geq 0}$.
In our networks, $h_t(u)$ is applied independently to each feature channel, with a separate learned time $t$ per-channel.
Learning the diffusion parameter is a key strength of our method, allowing the network to continuously optimize for spatial support ranging from purely local to totally global, and even choose different receptive fields for each feature (\figref{diffusion_viz}).
We thus sidestep challenges like manually choosing the support radius of a convolution, or sizes for a pooling hierarchy.

In the language of deep learning, diffusion can be viewed as a kind of smooth mean (average) pooling operation with many benefits: it has a geometrically-principled meaning, its support ranges from purely local to totally global via the choice of diffusion time, and it is differentiable with respect to diffusion time, allowing spatial support to be automatically optimized as a network parameter.

\paragraph{A note on generality.}

Remarkably, eschewing traditional representations of convolutions in favor of diffusion does not reduce the expressive power of our networks. This is supported by the following theoretical result (that we prove in \appref{generality}), which shows that radially-symmetric convolutions are contained in the function space defined by diffusion followed by a pointwise map:

\begin{lemma}[Inclusion of radially-symmetric convolutions]
    \label{thm:generality}
    For a signal $u: \mathbb{R}^2 \to \mathbb{R}$, let $U_r(p) : \mathbb{R}_{\geq 0} \to \mathbb{R}$ denote the integral of $u$ along the $r$-sphere at $p$, and let $u_t(p) : \mathbb{R}_{\geq 0} \to \mathbb{R}$ denote the signal value at \(p\) after diffusion for time \(t\).
    Then there exists a function transform $\mathcal{T}$ which recovers $U_r(p)$ from $u_t(p)$
    $$
    U_r(p) = \mathcal{T}[u_t(p)](r).
    $$
    Thus convolution with a radial kernel $\alpha: \RR_{\geq 0} \to \RR$ is given by
    $$
    (u  * \alpha)(p) =
    \int_{\mathbb{R}^2} \alpha(|q-p|) u_0(q)\ dq =
    \int_{0}^{\infty} \alpha(r) \mathcal{T}[u_t(p)](r)\ dr,
    $$
    which is a pointwise operation at $p$ on the diffused values $u_t(p)$.
 \end{lemma}

This fact is significant because it suggests that simple and robust diffusion can, without loss of generality, be used to replace complicated operations such as radial geodesic convolution.

Importantly, we will also extend our architecture beyond radially-symmetric filters by incorporating gradient features (\secref{spatial_gradients}).

\subsection{Computing Diffusion}
\label{sec:ComputingDiffusion}

Many numerical schemes could potentially be used to evaluate the diffusion layer $h_t(u)$, from direct solvers~\cite{chen2008algorithm} to hierarchical schemes~\cite{vaxman2010multi,liu2021surface}.
In particular, we seek schemes which are efficient as well as differentiable, to enable network training.
Here we describe two simple methods considered in our experiments.
The first scheme we consider is an implicit timestep, which is straightforward but requires solving large sparse linear systems, and the second is spectral expansion, which uses only efficient dense arithmetic at evaluation time but requires some modest precomputation.
Both are easily implemented using common numerical libraries, and we observe that networks trained with either approach have similar accuracy.
Efficiency is evaluated in \secref{performance}; we generally recommend spectral acceleration.

\subsubsection{Direct Implicit Timestep}

Perhaps the simplest effective approach to simulate diffusion is a single implicit Euler timestep
\begin{equation}
\label{eq:implicit_diffusion}
  h_t(u) := (M + tL)^{-1} M u
\end{equation}
which amounts to solving a (sparse) linear system for each diffusion operation.
Using an implicit backward timestep rather than an explicit forward timestep is crucial as it makes the scheme stable, allows global support, and yields a reasonable approximation of diffusion after just one step.
Solving linear systems (including derivative computation) is supported in modern learning software frameworks, allowing \eqref{implicit_diffusion} to implement a learnable diffusion block.
However, this amounts to solving a distinct large linear system for each channel, and GPU-based computation may fall back to solving dense linear systems, which means that direct implicit timesteps may not scale well to large problems.

\subsubsection{Spectral Acceleration}
\label{sec:SpectralAcceleration}

An alternate approach is to leverage a closed-form expression for diffusion in the basis of low-frequency Laplacian eigenfunctions~\cite{vallet2008spectral,zhang2010spectral}.
Once the eigenbasis has been precomputed, diffusion can then be evaluated for any time $t$ via elementwise exponentiation.
Truncating diffusion to a low-frequency basis incurs some approximation error, but we find that this approximation has little effect on our method (\figref{eig_sweep_plot}), perhaps because diffusion quickly damps high-frequency content regardless.

\setlength{\columnsep}{1em}
\setlength{\intextsep}{0em}
\begin{wrapfigure}{r}{110pt}
   \vspace{-0.1\baselineskip}
   \includegraphics[width=110pt]{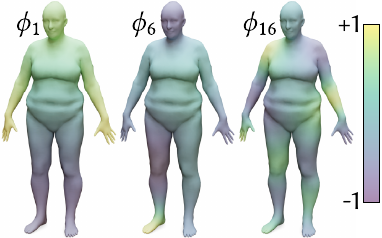}
   \vspace*{-1.5\baselineskip}
\end{wrapfigure}
For weak Laplace matrix $L$ and mass matrix $M$, the eigenvectors \(\phi_i \in \mathbb{R}^V\) are solutions to:

\begin{minipage}{4cm}
\begin{equation}
L \phi_i = \lambda_i M \phi_i,
\end{equation}
\end{minipage}\vspace{0.9em}

\noindent corresponding to the first \(k\) smallest-magnitude eigenvalues \(\lambda_1, \ldots, \lambda_k\). We normalize them so that $\phi_i^T M \phi_i = 1$.
These eigenvectors are easily precomputed for each shape of interest via standard numerical packages~\cite{lehoucq1998arpack}; the inset figure shows several example functions \(\phi_i\) for a surface of a human shape.

Let $\Phi := [\phi_i] \in \mathbb{R}^{V \times k}$ be the stacked matrix of eigenvectors, which form an orthonormal basis with respect to $M$. We can then project any scalar function $u$ to obtain its coefficients $c$ in the spectral basis via $c \gets \Phi^T M u$, and recover values at vertices as $u \gets \Phi c$.
Conveniently, diffusion for time $t$ is easily expressed as an elementwise scaling of spectral coefficients according to $c_i \gets e^{-\lambda_i t} c_i$.
The diffusion layer $h_t(u)$ is then evaluated by projecting on to the spectral basis, evaluating pointwise diffusion, and projecting back
\begin{equation}
  \label{eq:DiffusionLayer}
  h_t(u) := \Phi \begin{bmatrix}e^{-\lambda_0 t} \\ e^{-\lambda_1 t} \\ \dots \end{bmatrix} \odot (\Phi^T M u)
\end{equation}
where $\odot$ denotes the Hadamard (elementwise) product.
This operation is efficiently evaluated using dense linear algebra operations like elementwise exponentiation and matrix multiplication, and is easily differentiable with respect to $u$ and $t$.

\paragraph{Remarks}
We emphasize that DiffusionNet can still learn high-frequency outputs despite the use of a low-frequency basis (\eg{}, in \secref{sampling_invariance}).
Intuitively, diffusion is used for communication across points, for which a low-frequency approximation is typically sufficient, while MLPs and gradient features learn high-frequency features as needed for a task.
Additionally, we note that DiffusionNet is \emph{not} a spectral learning method---spectral coefficients are never used to represent filters or latent data, and thus no issues arise due to differing eigenbases on different shapes.
Spectral acceleration is merely one possible numerical scheme to compute diffusion.

\subsection{Spatial Gradient Features}
\label{sec:spatial_gradients}

\begin{figure}
\begin{center}
\end{center}
    \includegraphics{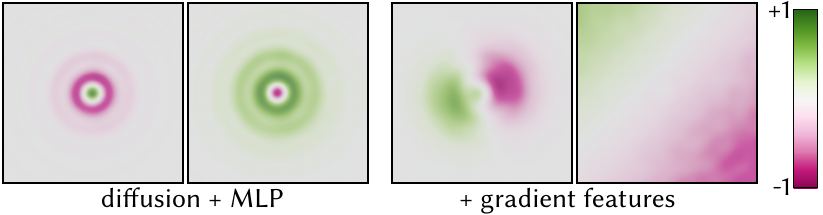}
    \caption{
      Diffusion followed by an MLP enables the network to learn radially-symmetric filters (\figloc{left}); introducing gradient features expands the space to include directional filters (\figloc{right}), while remaining invariant to the choice of local tangent basis.
      Here, we take a DiffusionNet block trained for segmentation and visualize the learned filter via channels of a normalized signal which maximizes the block output at the center point.
    \label{fig:filters_with_grad}}
\end{figure}

Our learned diffusion layer enables propagation of information across different points on a shape, but it supports only radially-symmetric filters about a point.
The last building block in our method enables a larger space of filters by computing additional features from the spatial gradients of signal values at vertices (\figref{filters_with_grad}).
Specifically, we construct features from the inner products between pairs of feature gradients at each vertex, after applying a learned scaling or rotation.

\paragraph{Evaluating gradients.}
We will express the spatial gradient of a scalar function on a surface as a 2D vector in the tangent space of each vertex.
These gradients can be evaluated by a standard procedure, choosing a normal vector at each vertex (given as input or locally approximated), then projecting neighbors into the tangent plane---either 1-ring neighbors on a mesh, or $m$-nearest neighbors in a point cloud.
The gradient is then computed in the tangent plane via least-squares approximation of the function values at neighboring points (see \citet{mukherjee2006estimation} for analysis).
These gradient operators at each vertex can be assembled into a sparse matrix $G \in \mathbb{C}^{V \times V}$, which is applied to a vector $u$ of real values at vertices to produce gradient tangent vectors at each vertex.
This matrix does not depend on the features, and can be precomputed once for each shape.
We use complex numbers as a convenient notation for tangent vectors, in an arbitrary reference basis in the tangent plane of each point (as in \cite{knoppel2013globally,Sharp:2019:VHM}, \etc{}).
If the normals are consistently oriented, then the imaginary axis is chosen to form a right-handed basis in 3D with respect to the outward surface normal.

\begin{figure}[b]
    \vspace*{-1em}
    \includegraphics{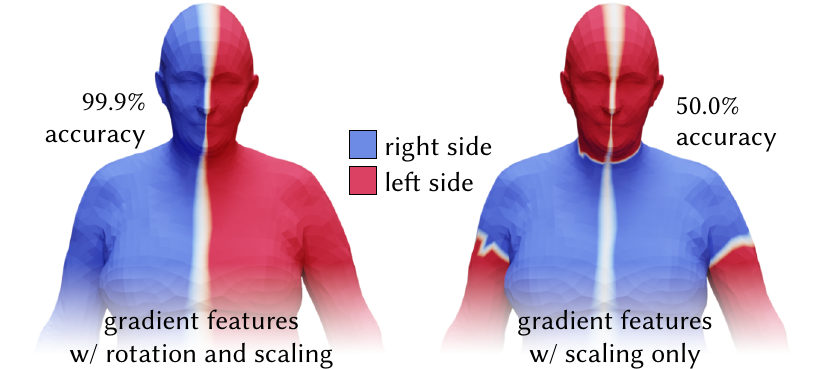}
    \caption{
    In a synthetic experiment, we demonstrate how our networks can successfully segment the left and right sides of bilaterally symmetric models even in a purely-intrinsic formulation (\emph{left}), because rotation in the tangent space for gradient features encodes a notion of orientation.
    Replacing this rotation with scaling (i.e., using a real matrix $A$ in \eqref{gradients}) removes the sensitivity to shape orientation (\emph{right}), but also avoids the need for consistent outward normals.
    See \appref{details} for details.
    \label{fig:leftright_segmentation}}
\end{figure}

\pagebreak
\paragraph{Learned pairwise products.}

Equipped with per-vertex spatial gradients of each channel, we learn informative scalar features by evaluating an inner product between pairs of feature gradients at each vertex, after a learned linear transformation.
Inner products are invariant to rotations of the coordinate system, so these features are invariant to the choice of tangent basis at vertices, as expected.
Putting it all together, given a collection of $D$ scalar feature channels, for each channel $u$ we first construct its spatial gradient as $z_u \in \mathbb{C}^{V}$, a vector of local 2D gradients per-vertex
\begin{equation}
z_u := G u
\end{equation}
then at each vertex $v$, we stack the local gradients of all channels to form $w_v \in \mathbb{C}^{D}$ and obtain real-valued features $g_v \in \mathbb{R}^D$ as
\begin{equation}
\label{eq:gradients}
g_v := \textrm{tanh}(\textrm{Re} (\overline{w}_v \odot A w_v))
\end{equation}
where $A$ is a learned square $D \times D$ matrix, and taking the real part $\textrm{Re}$ after a complex conjugate $\overline{w}_v$ is again just a notational convenience for dot products between pairs of 2D vectors.
This means the $i^\text{th}$ entry of the output at vertex $v$ is given by the dot product $g_v(i) = \textrm{tanh}\big(\textrm{Re} \left\{\sum_{j=1}^D \overline{w}_v(i) A_{ij} w_v(j) \right\} \big)$, so that each inner product is scaled by a learned coefficient $A_{ij}$.
The outer $\textrm{tanh}(\cdot)$ nonlinearity is not fundamental, but we find that it stabilizes training.

The choice of $A$ as a complex or real matrix has a subtle relationship with the orientation of the underlying surface.
Multiplying $w_v(j)$ by a \emph{complex} scalar both rotates and scales local gradient vectors before taking inner products (recall that complex multiplication can be interpreted as a rotation and scaling in the complex plane).
In contrast, real $A$ only allows scaling.
However, the direction of rotation (clockwise or counter-clockwise) depends on the choice of the outward normal, and hence on orientation---so surfaces with consistently oriented normals gain a richer representation by learning a complex matrix, whereas surfaces without consistent orientation (\eg{}, raw point clouds) should restrict to real $A$.

In \figref{leftright_segmentation}, a small synthetic experiment (detailed in \appref{details}) demonstrates how these learned rotations allow our method to disambiguate bilateral symmetry even in a purely intrinsic representation, a common challenge in non-rigid shape correspondence.

\begin{figure*}[t]
\begin{center}
   \includegraphics{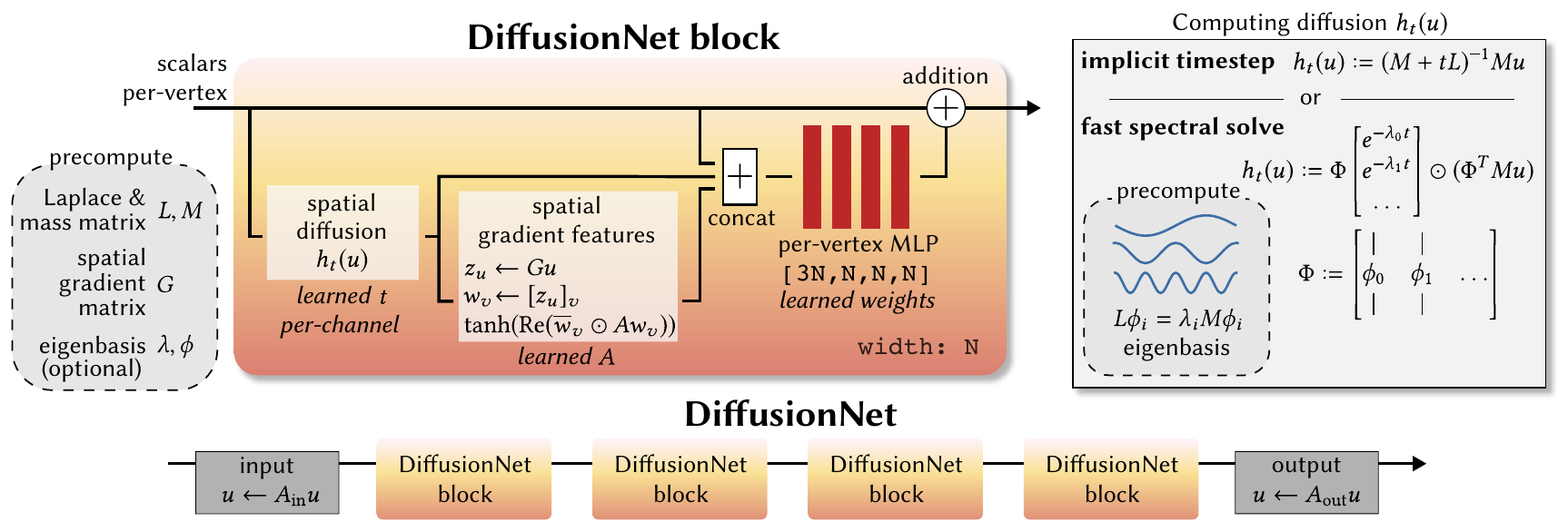}
\end{center}
\caption{
  We present DiffusionNet, a simple and effective architecture for learning on surfaces.
  It is composed of successive identical DiffusionNet blocks. Each block diffuses every feature for a learned time scale, forms spatial gradient features, and applies a \emph{spatially shared} pointwise MLP at each vertex in a mesh/point cloud/\etc{}
  These networks achieve state-of-the-art performance on surface learning tasks without any explicit surface convolutions or pooling hierarchies, in part because they automatically optimize for variable spatial support (see \eg{}, \figref{diffusion_viz}).
\label{fig:diffusionnet-architecture}
}
\end{figure*}

\section{DiffusionNet Architecture}
\label{sec:architecture}

The previous section establishes three ingredients for learning on surfaces: an MLP applied independently at each vertex to represent pointwise functions, learned diffusion for spatial communication, and spatial gradient features to model directional filters.
We combine these ingredients to construct \emph{DiffusionNet} (\figref{diffusionnet-architecture}), composed of several \emph{DiffusionNet blocks}.
This simple network operates on a fixed channel width $D$ of scalar values throughout, with each DiffusionNet block diffusing the features, constructing spatial gradient features, and feeding the result to an MLP.

We include residual connections to stabilize training~\cite{he2016deep}, as well as linear layers to convert to the expected input and output dimension.
When appropriate, results at the edges or faces of a mesh can be computed by averaging network outputs from the incident vertices, \eg{}, to segment the faces of a mesh.
Various activations can be appended to the end of the network based on the problem at hand, such as a softmax for segmentation, or a global mean followed by a softmax for classification; otherwise, this same architecture is used for all experiments.

Remarkably, we do not find it necessary to use any spatial convolutions or pooling hierarchies on surfaces---avoiding these potentially-complex operations helps keep DiffusionNet simple and robust.

\paragraph{Invariance.}
DiffusionNet is invariant to rigid motion of the underlying shapes as long as the input features remain unchanged, due to the intrinsic geometric nature of diffusion and spatial gradients.
The overall invariance then depends on the choice of input features.

\subsection{Input features}

DiffusionNet takes a vector of scalar values per-vertex as input features.
Here we consider two simple choices of features, others could easily be included when available.
Most directly, we simply use the raw 3D coordinates of a shape as input; rotation augmentation can be used to promote rigid invariance when inputs are not consistently aligned.
When rigid or even non-rigid invariance is desired, we instead use the Heat Kernel Signatures~\cite{sun2009concise} as input; these signatures are trivially computed from the spectral basis in \secref{SpectralAcceleration}.
Due to the intrinsic nature of our approach, with HKS as input, the networks are invariant to any orientation-preserving isometric deformation of the shape. Higher-order descriptors such as SHOT~\cite{tombari2010unique} seem unnecessary, and may be unstable under remeshing~\cite{Donati2020DeepGF}.

\section{Experiments and Analysis}
\label{sec:experiments}

The same network architecture achieves state-of-the-art results across many tasks, and more importantly offers new and valuable capabilities.
See the appendix for additional analyses.

\paragraph{Setup.}
We use the same basic 4-block DiffusionNet architecture and training procedure for all tasks,  varying the network size from a small 32-width ($30$k parameter) to a large 256-width ($1.8$M parameter) DiffusionNet according to the scale of the problem.
The shape of the first and last linear layers is adapted to the input and output dimension for the problem.
MLPs use ReLU activations and optionally dropout after intermediate linear layers.
We let ``xyz'' and ``hks'' denote networks with positions and heat kernel signatures as input, respectively.
All inputs are centered and scaled to be contained in a unit sphere, and heat kernel signatures are sampled at 16 values of $t$ logarithmically spaced on $[0.01, 1]$.
We do not use any data augmentation, except random rotations in tasks where positions are used as features yet a rotation-invariant network is desired.

We fit DiffusionNet using the ADAM optimizer with an initial learning rate of $0.001$ and a batch size of 1, training for $200$ epochs and decaying the learning rate by a factor of $0.5$ every $50$ epochs.
Cross-entropy loss is used for labelling problems.
Spectral acceleration is used to evaluate diffusion except where noted, truncated to a $k=128$ eigenbasis.
On point clouds, $30$ nearest-neighbors are used to assemble matrices.
Test accuracies are measured after the last epoch of training.

\paragraph{Implementation details.}
Precomputation to assemble matrices and compute the Laplacian eigenbasis for spectral acceleration is performed once as a preprocess on the CPU using SciPy~\cite{2020SciPy-NMeth, lehoucq1998arpack}.
Networks are implemented in PyTorch~\cite{paszke2019pytorch} and evaluated on a single GPU with standard backpropagation.
Performance is discussed in \secref{performance}; we find that DiffusionNet is very efficient and scalable compared to recent mesh-based learning methods.
Code and reproducible experiments are available at \href{https://github.com/nmwsharp/diffusion-net}{\color{blue}github.com/nmwsharp/diffusion-net}.

\begin{table}[]
\centering

\caption{
    DiffusionNet achieves nearly-perfect accuracy classifying 30-class SHREC11~\cite{lian2011shape} while training on just 10 samples per class.
    Results marked by $^\dagger$ are trained and tested on simplified models.
    \label{tab:shrec11-class}
}

\begin{tabular}{@{}lr@{}}
\toprule
\textbf{Method}     & \textbf{Accuracy} \\ \midrule
GWCNN \cite{ezuz2017gwcnn} &     90.3\%     \\
MeshCNN$^\dagger$ \cite{hanocka2019meshcnn}   & 91.0\%   \\
HSN$^\dagger$ \cite{wiersma2020cnns} & 96.1\% \\
MeshWalker$^\dagger$ \cite{lahav2020meshwalker} & 97.1\% \\
PD-MeshNet$^\dagger$ \cite{milano2020primaldual} & 99.1\% \\
HodgeNet$^\dagger$ \cite{smirnov2021hodgenet} & 94.7\% \\
FC$^\dagger$ \cite{mitchel2021field} & 99.2\% \\
DiffusionNet - xyz$^\dagger$ & {99.4\%}    \\
DiffusionNet - xyz & {99.0\%}    \\
DiffusionNet - hks$^\dagger$ & {99.5\%}    \\
DiffusionNet - hks & {\bftab 99.7\%}    \\
\bottomrule
\end{tabular}
\vspace*{-1em}
\end{table}

\subsection{Classification}
\label{sec:classification}

We first apply DiffusionNet to classify meshes in the SHREC-11 dataset~\cite{lian2011shape}, which has 30 categories of 20 shapes each.
We demonstrate that DiffusionNet learns successfully even in the presence of limited data.
As in the other cited results, we train on just 10 samples per class; our results are averaged over 10 trials of the experiment with random training splits.
We fit a cross-entropy loss with a label smoothing factor of $0.2$ (see discussion in \cite{goyal2021revisiting}).
We use a 32-width DiffusionNet for hks features, and 64-width DiffusionNet for xyz features with rotation augmentation.
DiffusionNet achieves the highest reported accuracy when applied directly on the original dataset models, or to the simplified models common in recent mesh-based learning work (\tabref{shrec11-class}).

\subsection{Segmentation}
\label{sec:segmentation}

\begin{figure}
\begin{center}
\end{center}
    \includegraphics{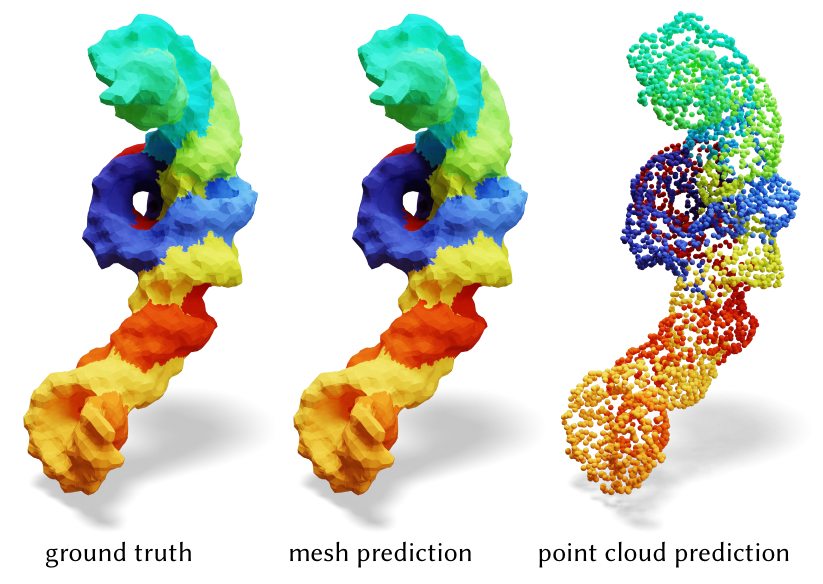}
    \caption{Segmenting RNA molecules with our method achieves accurate results when applied either directly to meshes or to sampled point clouds.
    \label{fig:rna_examples}}
\vspace{3em}
\end{figure}

\paragraph{Molecular segmentation.}
We evaluate a 128-width DiffusionNet on the task of segmenting RNA molecules into functional components, using the dataset introduced by \citet{poulenard2019effective}.
This dataset consists of 640 RNA surface meshes of about $15$k vertices each extracted from the Protein Data Bank~\cite{berman2000protein}, labelled at each vertex according to $259$ atomic categories, with a random 80/20 train-test split.
We learn these labels directly on the raw meshes, as well as on point clouds of 4096 uniformly-sampled points as in past work \citet{poulenard2019effective}.
For comparison, we cite point cloud results reported in ~\cite{poulenard2019effective}, and additionally train methods related to ours, SplineCNN~\cite{fey2018splinecnn} and a Dirac Surface Network~\cite{kostrikov2018surface} on meshes.
We also attempted to train MeshCNN~\cite{hanocka2019meshcnn} and HSN~\cite{wiersma2020cnns}, but found the former prohibitively expensive, while the latter could not successfully preprocess the data.
Our method achieves state-of-the-art accuracy on both the mesh and point cloud variants of the problem (\tabref{rna-segmentation}, \figref{rna_examples}).
Learning directly on the mesh yields greater accuracy, perhaps because no information is lost when sampling a point cloud, and surface structure is preserved.

\begin{table}[]
\centering
\caption{
    Accuracy of various mesh and point cloud schemes for RNA segmentation.
    DiffusionNet achieves state-of-the-art results, in part because it can be applied directly to the raw meshes.
    ``xyz'' and ``hks'' denote networks taking raw coordinates or heat kernel signatures as input, respectively.
    \label{tab:rna-segmentation}
}
\begin{tabular}{@{}rlr@{}}
\cmidrule[\heavyrulewidth]{2-3}
 & \textbf{Method}     & \textbf{Accuracy} \\
\cmidrule{2-3}
\ldelim\{{5}{.73cm}[cloud] \hspace{0.5em}
 & PointNet++ \cite{qi2017pointnet} & 74.4\% \\
 & PCNN ~\cite{atzmon2018pcnn} & 78.0\% \\
 & SPHNet \cite{poulenard2019effective} & 80.1\% \\
 & DiffusionNet - hks & {84.0\%}    \\
 & DiffusionNet - xyz & {\bftab 85.0\%}    \\
\cmidrule{2-3}
  \ldelim\{{4}{.7cm}[mesh] \hspace{0.5em}  &  SplineCNN \cite{fey2018splinecnn} & 53.6\%    \\
 & SurfaceNetworks \cite{kostrikov2018surface} & 88.5\% \\
 & DiffusionNet - hks & 91.0\%    \\
 & DiffusionNet - xyz & {\bftab 91.5}\%    \\
\cmidrule[\heavyrulewidth]{2-3}
\end{tabular}
\end{table}

\paragraph{Human segmentation.}

We train a 128-width DiffusionNet with dropout to segment the human body parts on the composite dataset of~\cite{maron2017convolutional}, containing models from several other human shape datasets~\cite{bogo2014faust, anguelov2005scape, adobe2016mixamo, vlasic2008articulated, giorgi2007watertight}.
Additionally, we cite a variety of reported results from other approaches on this task, as reported by the respective original authors and by \citet{wiersma2020cnns}.
For clarity, we distinguish between variants of this task in past work which used simplified meshes and soft ground truth; more details in \appref{details}.
Our model is quite effective using both rotation-augmented raw coordinates or heat kernel signatures as input (\tabref{human-segmentation}).

\begin{table}[]
\centering
\caption{
  Human part segmentation on the dataset of \citet{maron2017convolutional}.
  \emph{xyz} and \emph{hks} denote DiffusionNet with positions or heat kernel signatures as input, respectively.
  The $^\dagger$ rows use simplified inputs with a soft ground truth at edges, and the $^\ddagger$ rows use simplified inputs with hard ground truth at faces; details in \appref{details}.
  \label{tab:human-segmentation}
}
\begin{tabular}{@{}rlr@{}}
\cmidrule[\heavyrulewidth]{2-3}
& \textbf{Method}     & \textbf{Accuracy} \\
\cmidrule{2-3}
& GCNN~\cite{masci2015geodesic} & 86.4\% \\
& ACNN~\cite{boscaini2016learning} & 83.7\% \\
& Toric Cover~\cite{maron2017convolutional} & 88.0\%  \\
& PointNet++~\cite{qi2017pointnet} & 90.8\% \\
& MDGCNN [Poulenard et al. 2018] & 88.6\% \\
& DGCNN~\cite{wang2019dgcnn} & 89.7\% \\
& SNGC~\cite{haim2019surface} & 91.0\% \\
& HSN~\cite{wiersma2020cnns}  & 91.1\% \\
& MeshWalker~\cite{lahav2020meshwalker} & {\bftab 92.7\% }\\
& CGConv~\cite{yang2021continuous}  & 89.9\% \\
& FC~\cite{mitchel2021field} & 92.5\% \\
& DiffusionNet - xyz & 90.6\%    \\
& DiffusionNet - hks & {91.7}\%     \\
\cmidrule{2-3}
\ldelim\{{4}{.1cm}[$\dagger$] \hspace{.2em}
& MeshCNN~\cite{hanocka2019meshcnn} & 92.3\%                        \\
& MeshWalker \cite{lahav2020meshwalker} & 94.8\%                    \\
& DiffusionNet - xyz & {\bftab  95.5}\%                             \\
& DiffusionNet - hks & {\bftab  95.5}\%                              \\
\cmidrule{2-3}
\ldelim\{{4}{.1cm}[$\ddagger$] \hspace{.2em}
& PD-MeshNet \cite{milano2020primaldual}    & 85.6\%                   \\
& HodgeNet \cite{smirnov2021hodgenet}       & 85.0\%                                 \\
& DiffusionNet - xyz & {\bftab  90.3}\%                     \\
& DiffusionNet - hks & {\bftab  90.8}\%                     \\
\cmidrule[\heavyrulewidth]{2-3}
\end{tabular}
\end{table}

\vspace{2em}

\subsection{Functional Correspondence}
\label{sec:functional_correspondence}

\begin{figure*}
\begin{center}
    \includegraphics{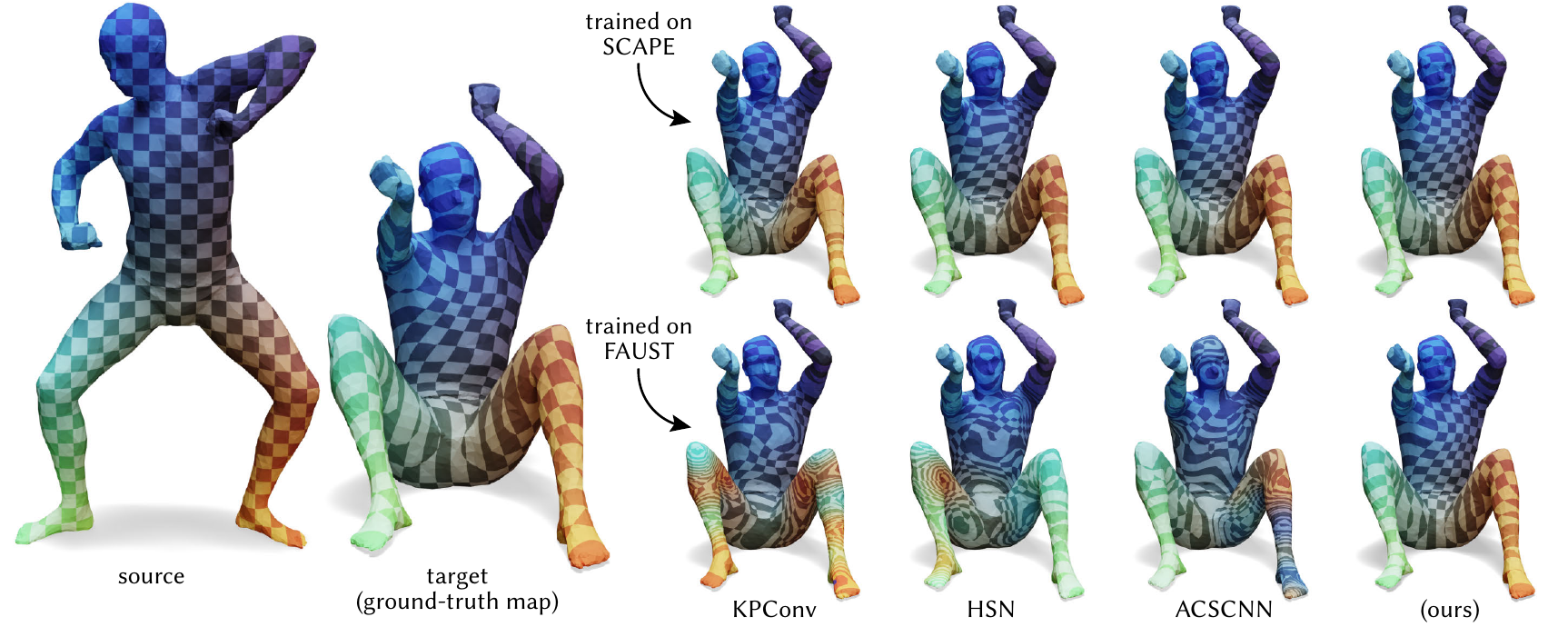}
    \caption{
      DiffusionNet is highly effective as a feature extractor for nonrigid correspondence via functional maps, shown here in the supervised setting (\secref{functional_correspondence}).
      Correspondences are visualized by transferring a texture through the map.
      All methods yield a visually plausible solution when trained on the same dataset as the query pair (SCAPE, \emph{top row}), but only DiffusionNet yields good results when generalizing after training on a different dataset (FAUST, \emph{bottom row}).
      \label{fig:functional_textures}
    }
\end{center}
\end{figure*}

Functional maps compute a correspondence between a pair of shapes by finding a linear transformation between spectral bases, aligning some set of input features \cite{ovsjanikov2012functional}.
Recent work has shown that learned features can improve performance, \eg{}, ~\citet{litany2017deep,Donati2020DeepGF}.
Here we demonstrate that using DiffusionNet as a feature extractor outperforms other recent approaches, yielding to state-of-the-art correspondence results in both the supervised and weakly-supervised variants of the problem.
We emphasize that the spectral representation in functional maps is unrelated to the spectral acceleration from \secref{SpectralAcceleration}, which is merely a scheme for evaluating diffusion; DiffusionNet itself does not learn in the spectral domain.

\begin{table}[]
\centering
\caption{
  Our approach yields state-of-the-art correspondences when used as a feature extractor for deep functional maps, both in the supervised (\figloc{top}, as in \citet{Donati2020DeepGF}) and the weakly supervised setting (\figloc{bottom}, as in \citet{sharma2020weakly}).
  The dotted rows apply ZoomOut post-processing to the previous result~\cite{melzi2019zoomout}.
  X on Y means train on X and test on Y.
  Values are mean geodesic error $\times 100$ on unit-area shapes.
  \label{tab:deep_functional_maps}
}
\setlength\tabcolsep{3pt}
\begin{tabular}{@{}lcccc@{}}
\toprule
\textbf{Method / Dataset}     & \textbf{F}AUST  & \textbf{S}CAPE  & \textbf{F}\! on\! \textbf{S}  & \textbf{S}\! on\! \textbf{F} \\ \midrule
KPConv \cite{thomas2019kpconv} & 3.1 & 4.4 & 11.0 & 6.0 \\
KPConv - hks  & 2.9 & 3.3 & 10.6 & 5.5 \\
HSN \cite{wiersma2020cnns}   & 3.3 & 3.5 & 25.4 & 16.7 \\
ACSCNN \cite{li2020shape}   & \textbf{2.7} & 3.2 & 8.4 & 6.0 \\
DiffusionNet - hks & \textbf{2.7} & \textbf{3.0} & 3.8 & \textbf{3.0} \\
DiffusionNet - xyz & \textbf{2.7} & \textbf{3.0} & \textbf{3.3} & \textbf{3.0} \\
\hdashline
\quad + ZoomOut & 1.9 & 2.4 & 2.4 & 1.9 \\
\bottomrule
WSupFMNet & \textbf{3.3} & 7.3 & 11.7 & 6.2 \\
WSupFMNet + DiffusionNet - xyz & 3.8 & \textbf{4.4} & \textbf{4.8} & \textbf{3.6} \\
\hdashline
\quad + ZoomOut & 1.9 & 2.6 & 2.7 & 1.9 \\
\bottomrule
\end{tabular}
\end{table}

Our experiments follow the setup of \citet{Donati2020DeepGF}, training and evaluating on both SCAPE~\cite{anguelov2005scape,ren2018continuous}, and FAUST~\cite{bogo2014faust} including training on one dataset and evaluating on the other.
In the \emph{supervised} setting we fit dataset-provided correspondences, and we generate rigid-invariant models by randomly rotating all inputs for training and testing.
For the \emph{weakly-supervised} setting, we use the dataset and losses advocated in~\cite{sharma2020weakly}, where rigid alignment of the input is used as weak supervision, without known correspondences.
In all cases we extract point-to-point maps between test shapes and evaluate them against ground truth dense correspondences; for simplicity we compare all methods without post-processing the maps, though we also report accuracies after postprocessing with ZoomOut for our method~\cite{melzi2019zoomout}.
In addition to citing results obtained using the KPConv feature extractor ~\cite{thomas2019kpconv} by \citet{Donati2020DeepGF}, we also train HSN~\cite{wiersma2020cnns}, ACSCNN~\cite{li2020shape}, and our own 128-width DiffusionNet with dropout.
We also tried MeshCNN~\cite{hanocka2019meshcnn}, but it proved to be prohibitively expensive at $14$hr per epoch.

As shown in \tabref{deep_functional_maps}, DiffusionNet yields state-of-the-art results for non-rigid shape correspondence in the both the supervised and weakly-supervised settings, especially when transferring between datasets.
One might wonder whether the improvement in this task truly stems from our DiffusionNet architecture, or from the use of HKS features. As shown in the same table, training KPConv with HKS as features shows DiffusionNet yields significant improvements regardless.
\figref{functional_textures} visualizes the resulting correspondences on a challenging test pair, where only DiffusionNet achieves a high-quality correspondence when generalizing after training on a different dataset.

\pagebreak
\subsection{Discretization Agnostic Learning}
\label{sec:sampling_invariance}

\begin{figure}
\begin{center}
\end{center}
    \includegraphics{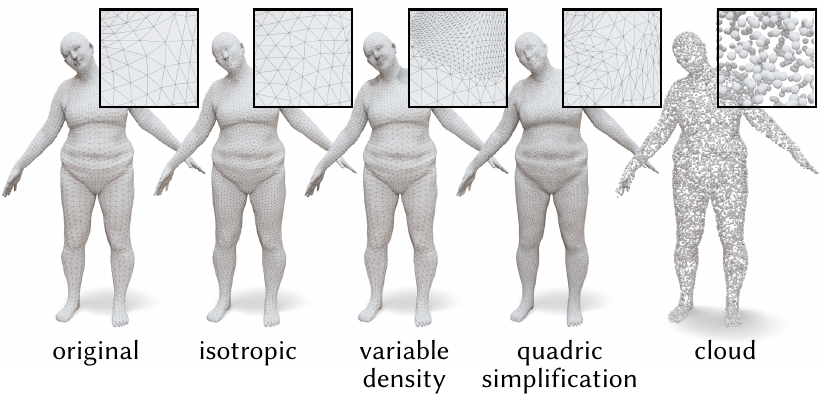}
    \caption{
      Examples of the remeshed FAUST test dataset used in \tabref{sampling_invariance}.
    \label{fig:sampling_invariance_viz}}
\end{figure}

\begin{table}[]
\centering
\caption{
  DiffusionNet automatically retains highly accurate results under changes in meshing and sampling, while many other approaches overfit to mesh connectivity.
  Here we give correspondence errors on our remeshed FAUST dataset after training on template meshes, measured in mean geodesic distance $\times 100$ after normalizing by the geodesic diameter.
  \label{tab:sampling_invariance}
}
\begin{tabular}{@{}lrR{2.75em}R{2.75em}R{2.75em}R{2.75em}@{}}
\toprule
& & \multicolumn{4}{c}{remeshed/sampled variants} \\
\cmidrule{3-6}

\textbf{Method}     & \textbf{orig} & \textbf{iso} & \textbf{dense} & \textbf{qes} & \textbf{cloud} \\ \midrule
ACSCNN              & {\bftab 0.05} & 35.29 & 19.09 & 41.15 & N/A\\
SplineCNN           & 3.51 & 31.09 & 27.95 & 40.43 & N/A\\
HSN                 & 9.57 & 20.01 & 24.84 & 25.40 & N/A\\
PointNet (vertices) & 3.83 & 2.92 & 3.04 & 2.67 & {\bftab 2.60}\\
PointNet (sampled)  & 9.99 & 4.25 & 7.84 & 4.44 & 4.13\\
DGCNN (vertices)    & 2.44 & 14.58 & 15.72 & 27.00 & 32.13\\
DGCNN (sampled)     & 6.52 & 4.30 & 5.57 & 3.61 & 2.66\\
DiffusionNet        & 0.33 & {\bftab 0.68} & {\bftab 0.62} & {\bftab 0.82} & {\bftab 2.59}\\

\bottomrule
\end{tabular}
\end{table}

\begin{figure}
\begin{center}
    \includegraphics{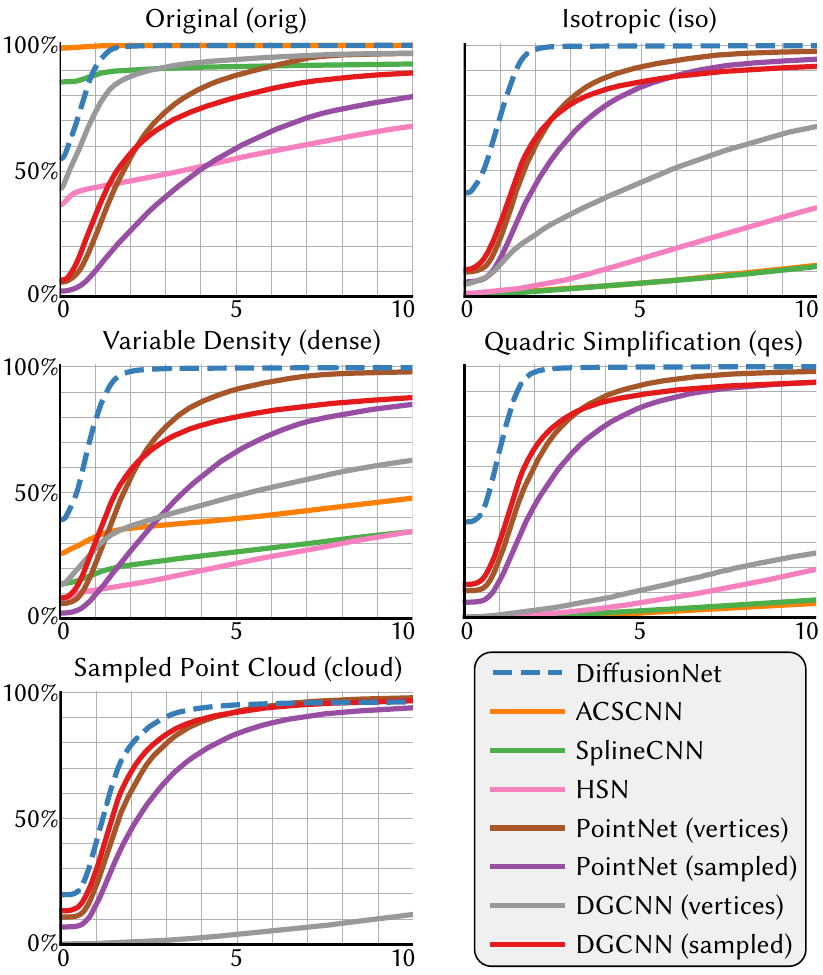}
    \caption{
      Accuracy curves for vertex-labelling correspondence on the FAUST dataset, as in \tabref{sampling_invariance}.
      The first plot gives accuracy on the original test meshes, and the subsequent plots denote testing on our remeshed variants of the test set.
      For each plot, the x-axis is the geodesic error $\times 100$ after normalizing by geodesic diameter, and the y-axis is the percent of predicted correspondences within that error.
      \label{fig:eval_correspondence}
    }
\end{center}
\end{figure}

A key benefit of DiffusionNet compared to many past approaches is that its outputs are robust to changes in the discretization of the input (\eg{}, different meshes of the same shape, or a mesh \vs{} a point cloud).
This property is essential for practical applications where meshes for inference are likely to be tessellated differently from the training set, \etc{}
Other invariants (\eg{}, rigid invariance) can be encouraged via data augmentation, but for discretization it is impractical to generate augmented inputs across all possible sampling patterns.
Below, we demonstrate that  DiffusionNet generalizes quite well across discretizations without any special regularization or augmentation, especially compared to recent mesh-based methods.

We study discretization invariance using a popular formulation of the shape correspondence task on the FAUST human dataset~\cite{bogo2014faust}, where each vertex of a mesh is to be labelled with the corresponding vertex on a template mesh.
Importantly, these input meshes are already manually aligned templates, with exactly identical mesh connectivity.
Past work has achieved near-perfect accuracy in this problem setup (\eg{}, \cite{fey2018splinecnn, li2020shape}), however we suggest that these models primarily overfit to mesh graph structure. In contrast, DiffusionNet learns an accurate and general function of the shape itself, despite the synthetic setup.

To experimentally quantify this effect, we construct a version of the FAUST test set after remeshing with several strategies: \emph{orig} is the original test mesh, \emph{iso} is a uniform isotropic remeshing, \emph{dense} refines the mesh in randomly sampled regions, \emph{qes} first refines the meshes, then applies quadric error simplification~\cite{garland1997surface}, and \emph{cloud} is a point cloud with normals sampled from the surface (\figref{sampling_invariance_viz}).
Unrelated remeshings have appeared in~\cite{poulenard2018multi,wiersma2020cnns}, but the procedure therein left large regions of the mesh unchanged.
Ground truth for evaluation is defined via nearest-neighbor on the original test mesh.
We train on the 3D coordinates of the 80 standard FAUST registered meshes, but evaluate on the remeshed set to mimic the practical scenario where the training set contains meshes tessellated via some particular common strategy, yet the fitted model must be applied to totally different meshes encountered in the wild.
We also train several other methods---details in \appref{details}.

\tabref{sampling_invariance} and \figref{eval_correspondence} show how other mesh-based approaches degrade rapidly under remeshing; only DiffusionNet yields accurate correspondences which are largely stable under remeshing and resampling.
Some point-based methods avoid the dependence on connectivity, yet do not match the overall accuracy of the surface-based DiffusionNet.

\begin{table}[b]
\caption{
  Runtimes of DiffusionNet and other mesh-based methods across several different input mesh resolutions.
  Reported times are for one-time preprocessing (\textit{pre}), a training evaluation with derivatives (\textit{train}), and an inference evaluation (\textit{infer}), each on a single input of the specified size.
  Entries marked by ``---'' were infeasibly expensive in time or memory usage.
  Unlike many recent mesh-based learning methods, DiffusionNet easily trains directly on medium-sized inputs, and even scales to very large meshes.
  \label{tab:runtime}
  \vspace{-1em}
}
\begin{tabular}{@{}crlll@{}}
\toprule
                &  & \multicolumn{1}{c}{\textbf{small}} & \multicolumn{1}{c}{\textbf{medium}}  & \multicolumn{1}{c}{\textbf{large}}  \\
\textbf{Method} &  &  \multicolumn{1}{c}{752 vert} & \multicolumn{1}{c}{10k vert}  & \multicolumn{1}{c}{184k vert}  \\
\midrule
                           & \textit{pre:}    &  288ms  & 3.55sec   & 69.5sec \\
\textbf{DiffusionNet}      & \textit{train:}  &  19ms   & 25ms      & 379ms \\
(spectral)                 & \textit{infer:}  &  7ms    & 10ms      & 154ms \\
\midrule
                           & \textit{pre:}    &  104ms  & \nodata   & \nodata \\
\textbf{DiffusionNet}      & \textit{train:}  &  329ms  & \nodata   & \nodata \\
(direct)                   & \textit{infer:}  &  81ms   & \nodata   & \nodata \\
\midrule
                           & \textit{pre:}    &  85ms   & 1.13sec  & \nodata \\
\textbf{MeshCNN}           & \textit{train:}  &  269ms  & 2.97sec   & \nodata \\
\cite{hanocka2019meshcnn}  & \textit{infer:}  &  194ms  & 2.71sec   & \nodata \\
\midrule
                           & \textit{pre:}    &  905ms  & 162sec    & \nodata  \\
\textbf{HSN}               & \textit{train:}  &  188ms  & 1.08sec   & \nodata  \\
\cite{wiersma2020cnns}     & \textit{infer:}  &  68ms   & 389ms     & \nodata  \\
\midrule
                           & \textit{pre:}    &  \textit{n/a} &  \textit{n/a}      & \nodata \\
\textbf{HodgeNet}          & \textit{train:}  &  752ms  &  7.61sec  & \nodata \\
{[Smirnov et al. 2021]}    & \textit{infer:}  &  645ms  &  6.87sec  & \nodata \\
\bottomrule
\end{tabular}
\end{table}

\subsection{Transfer Across Representations}

Not only are the outputs of DiffusionNet consistent across remeshing and resampling of a shape, but furthermore the same network can be directly applied to different discrete representations.
The only geometric data required for DiffusionNet are the Laplacian, mass, and spatial gradient matrices, which are easily constructed for many representations.
Because all geometric operations in the network are defined in terms of these standard matrices, fitted network weights retain the same meaning across different representations.
This enables us to train on one representation and evaluate on another as seen in  \figref{generalization_teaser} and \secref{sampling_invariance}, \emph{cloud}, without any special treatment or fine-tuning.
In the future, DiffusionNet opens the door to heterogeneous training sets which intermingle mesh, point cloud, and other surface data from various sources.

\subsection{Efficiency and Robustness}
\label{sec:performance}

\paragraph{Runtime}
DiffusionNet requires only standard linear algebra operations for training and inference, and is thus straightforward and efficient on modern hardware.
As an example, DiffusionNet with spectral acceleration trains on the 14k-vertex RNA meshes (\secref{segmentation}) in 38ms per input and requires 2.2GB of GPU memory.  
Preprocessing is performed on the CPU once for each input; for these RNA meshes preprocessing takes 5.4sec and generates 12MB of data each, composed mainly of the Laplacian eigenbasis $\Phi$ for spectral acceleration.
\tabref{runtime} summarizes the runtime performance of DiffusionNet and several other recent methods on the human segmentation task (\secref{segmentation}), including preprocessing, training with gradient computation, and inference.
All timings are measured on a 24GB Titan RTX GPU and dual Xeon 5120 2.2Ghz CPUs.

\paragraph{Scaling}
Most significantly, DiffusionNet's efficiency enables direct learning on common mesh data without dramatic simplification, in contrast to other recent mesh-based schemes.
As an example, the segmentation meshes from \citet{maron2017convolutional} have up to 13k vertices, yet recent approaches simplify/downsample to roughly $1$k vertices for training, as shown in \figref{resolution}~\cite{hanocka2019meshcnn,wiersma2020cnns,lahav2020meshwalker, mitchel2021field}.
In contrast, our networks easily run at full resolution on this and other datasets, paving the way for adoption in practice and improving accuracy due to preserved details (\eg{}, in \tabref{rna-segmentation}).
We even demonstrate DiffusionNet on a large, 184k vertex raw scan mesh from FAUST---again no special treatment is needed (\tabref{runtime}, \figref{generalization_teaser}).

\paragraph{Robustness}
DiffusionNet is also very robust to poor-quality input data; diffusion is a stable smoothing operation, and our method does not require any complex geometry processing operations such as geodesic distance~\cite{masci2015geodesic}, edge collapse~\cite{hanocka2019meshcnn}, parallel transport~\cite{wiersma2020cnns}, or managing pooling hierarchies with upsampling/downsampling.
Even the gradient matrix $G$ is the result of a stable least-squares fit.
If desired, techniques like the intrinsic Delaunay Laplacian on meshes can be used to further increase robustness~\cite{bobenko2007discrete,sharp2020laplacian}, though we do not find it necessary in our experiments.
We demonstrate in \figref{generalization_teaser} that DiffusionNet can be applied directly to a low-quality, nonmanifold raw scan mesh without any issues.

\section{Conclusion}
\label{sec:conclusion}

We present a new approach for learning on surfaces that is built by using learned diffusion as the main network component, with spatial gradient features to inject directional information.
Our method is very efficient to train and evaluate, is robust to changes in sampling, and even generalizes across representations, in addition to achieving state-of-the-art results on a range of tasks.

\setlength{\columnsep}{1em}
\setlength{\intextsep}{0em}
\begin{wrapfigure}{r}{90pt}
   \vspace*{-1.0\baselineskip}
   \includegraphics[width=90pt]{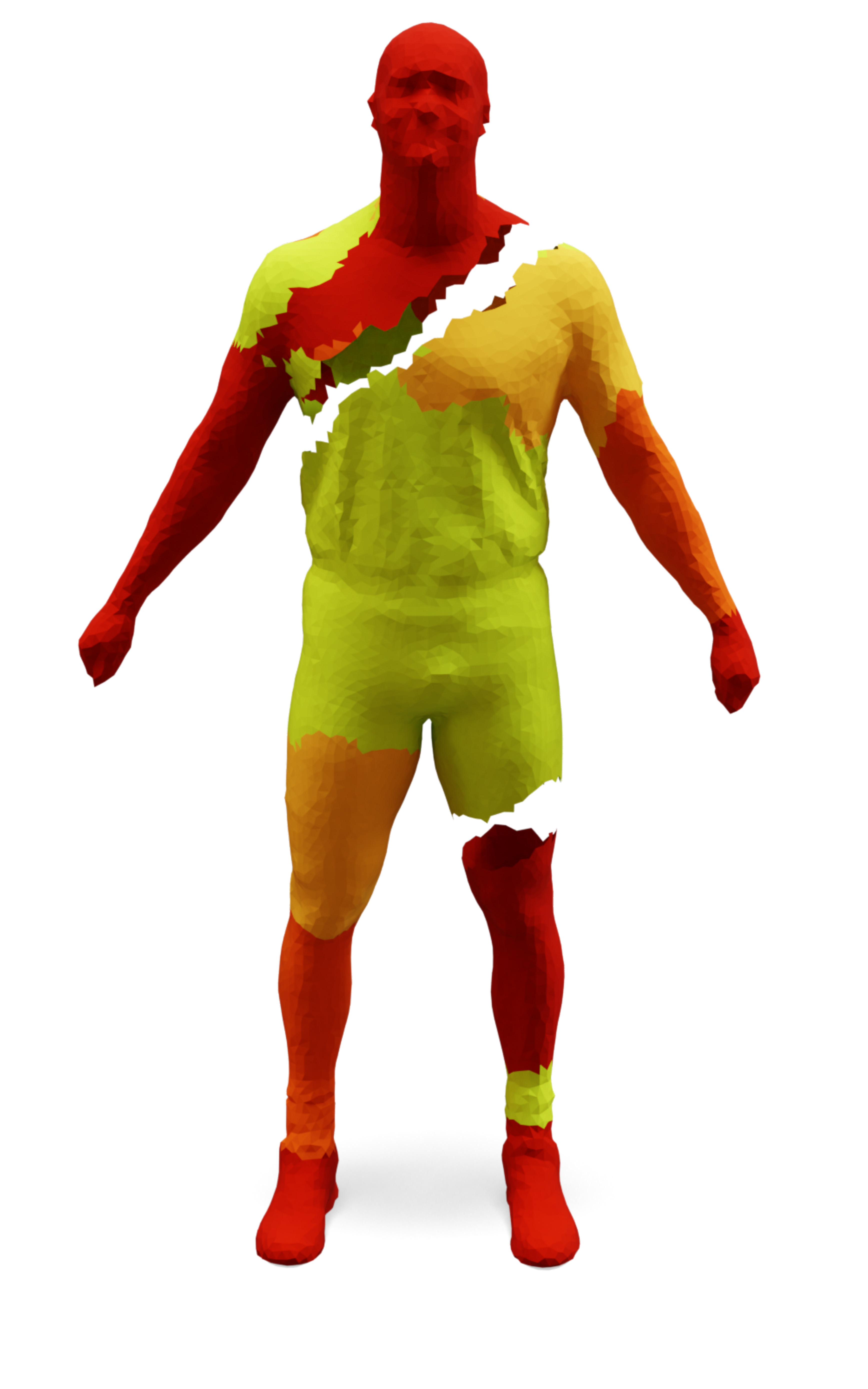}
   \vspace*{-2.5\baselineskip}
   \caption{Erroneous segmentation results due to disconnected components.\label{fig:limitation_cut}}
\end{wrapfigure}
\paragraph{Limitations}
DiffusionNet is designed to leverage the geometric structure of a surface; consequently it is not automatically robust to topological errors or outliers.
In fact, diffusion does not allow any communication at all between distinct components of a surface, leading to nonsensical outputs in the presence of spuriously disconnected components (see inset).
Subsequent work might mitigate the limitation by combining diffusion with other notions of communication, such as global pooling (\ala{}~\cite{qi2017pointnet}) or edge convolutions over latent nearest-neighbors~\cite{wang2019dgcnn}.

\pagebreak
Our networks are intentionally agnostic to local discretization, and thus may not be suited for tasks where one learns some property of the local discrete structure, such as denoising or mesh modification.
Finally, although our method discourages overfitting to mesh sampling (\secref{sampling_invariance}), it cannot guarantee to totally eliminate it, and we still observe a small drop in performance when transferring between representations---further investigation will seek to close this gap entirely.

\paragraph{Future work}

DiffusionNet can be applied to any surface representation for which a Laplacian matrix and spatial gradients can be constructed.
This opens the door to directly learning---and even transferring pretrained networks---on a wide variety of surface representations, from occupancy grids~\cite{caissard2019laplace} to subdivision surfaces~\cite{de2016subdivision}.
More broadly, DiffusionNet need not be restricted to explicit surfaces, and could easily be adapted to other geometric domains like volumetric meshes, curve networks, implicit level sets, depth maps, or images.
We believe that grounding geometric deep learning in the mathematically and computationally well-established diffusion operation will offer benefits across surface learning and beyond.

\section*{Acknowledgements}
The authors thank Adrien Poulenard and Ruben Wiersma for assistance configuring experiments, and Mark Gillespie for suggesting the mirrored training for \figref{leftright_segmentation}.
Parts of this work were generously supported by the Fields Institute for Research in Mathematical Sciences, the Vector Institute for AI, an NSF Graduate Research Fellowship, ERC Starting Grant No. 758800 (EXPROTEA) the ANR AI Chair AIGRETTE, a Packard Fellowship, NSF CAREER Award 1943123, and gifts from Activision Blizzard, Adobe, Disney, Facebook, and nTopology.

\bibliographystyle{ACM-Reference-Format}
\bibliography{DiffusionNet}

\appendix
\section{An Argument for Generality}
\label{app:generality} 

In \secref{learned_diffusion}, we propose diffusion at various learned timescales followed by a learned pointwise function as the essential components of our method.
Although this formulation clearly offers nonlocal support to the pointwise functions, it is not immediately clear how general the resulting function space is.
In particular, it is significant to show that this function space includes at least radially symmetric convolutions, a basic building block which has appeared widely in past work.
The treatment of radially symmetric convolutions arises because points on surfaces do not generally have canonical tangent coordinates, though it should be noted that recent work has since focused on expanding beyond symmetric filters, and our own method includes gradient features for precisely this purpose.
Lemma 1 states that, at least in the flat, continuous setting, this function space is sufficiently general to represent radially symmetric convolutions.
Here, we give a full version of this argument and some discussion.

Consider a scalar field $u: \mathbb{R}^2 \to \mathbb{R}$ in the plane.
Let $U_r(p): \mathbb{R}_{\geq 0} \to \mathbb{R}$ denote the integral of the field $u$ along the sphere with radius $r$ centered at $p$, \ie{} $U_r(p) = \int_{\partial B(p,r)} u(y) dy$. Recall that $u_t(p) : \mathbb{R}_{\geq 0} \to \mathbb{R}$ denotes the value of $u$ at $p$ after diffusion for time $t$.
We are interested in $U_r(p)$, because it will enable the evaluation of radially-symmetric convolutions against $u$.
The crux of our argument is to show that $U_r(p)$ can be recovered from $u_t(p)$, which we will formalize by showing the existence of a function transform 
\begin{equation*}
\mathcal{T}: (\mathbb{R}_{> 0} \to \mathbb{R}) \to (\mathbb{R}_{> 0} \to \mathbb{R})
\end{equation*}
such that
\begin{equation}
U_r(p) = \mathcal{T}[u_t(p)](r).
\end{equation}
The heat kernel solution for $u_t(p)$ is given by
\begin{equation}
\label{eq:heat_kernel_int}
u_t(p) = 
\int_{\mathbb{R}^2} u(q) \frac{1}{4 \pi t} e^{-\frac{|p-q|^2}{4t}} dq =
\int_{0}^\infty U_r(p) \frac{1}{4 \pi t} e^{-\frac{r^2}{4t}} dr,
\end{equation}
where the second equality moves to a radial integral, recalling that $U_r(p)$ is defined as the integral of $u$ along the sphere of radius $r$ at $p$.
Calculation verifies that this integral has the form of a Laplace transform of $U_r(p)$
\begin{equation}
u_t(p) = \frac{1}{4\pi t}\mathcal{L}[\frac{1}{2 \sqrt{r}}U_{\sqrt{r} }(p)](\frac{1}{4t}).
\end{equation}
The Laplace transform is injective~\cite{lerch1903point}, which allows us to consider the inverse transform
\begin{equation*}
U_r(p) = \mathcal{T}[u_t(p)](r).
\end{equation*}
And in fact, $\mathcal{T}$ will have the form of an inverse Laplace transform, up to reparameterization by $\frac{1}{t}$ and constant coefficients.  

Now that we have established the existence of $\mathcal{T}$, it is straightforward to evaluate a radially-symmetric convolution via a pointwise map applied to diffused values.
Convolution against any radially symmetric kernel $\alpha(r) : \mathbb{R}_{\geq 0} \to \mathbb{R}$ is given by
\begin{align}
  (u * \alpha)(p) &= \int_{\mathbb{R}^2} \alpha(|p-q|) u(q) dq \nonumber \\
          &= \int_{0}^\infty \alpha(r) U_r(p) dr \\
          &= \int_{0}^\infty \alpha(r) \mathcal{T}[u_t(p)](r) dr \nonumber.
\end{align}
In this sense, the function space defined by diffusion followed by a pointwise map contains the space of radially symmetric convolutions, completing our argument.

Extending this treatment from $\mathbb{R}^2$ to curved manifolds would require a deeper analysis, though the same essential properties hold for diffusion on surfaces.
Furthermore, we treat only the continuous setting above, rather than the discrete setting where pointwise maps are approximated via finite-dimensional MLPs, and diffusion is evaluated at a collection of times $t$.
More generally, it would be valuable to extend this analysis to formalize the stability properties of diffusion, \ala{} \cite{kostrikov2018surface,perlmutter2020scattering}.
Nonetheless, we consider this argument to be important evidence that diffusion followed by pointwise functions is an expressive function space, supported by the strong results of our method in practical experiments.

\section{Analysis}
\label{app:analysis}

\paragraph{Ablation}

\begin{table}[]
\centering
\caption{
    An ablation study, evaluated on the human segmentation task. Omitting any of the components of our method leads to a significant drop in performance.
    Manually fixing a non-optimal diffusion time also impairs performance---our learned procedure automatically optimizes a diffusion time for each channel.
    \label{tab:ablation}
}
\setlength\tabcolsep{3.5pt} 
\begin{tabular}{@{}lr@{}}
\toprule
\textbf{Ablation}     & \textbf{Accuracy}  \\ \midrule
no diffusion & 31.4 \% \\
fixed-time diffusion $t=0.1$ & 89.1 \% \\ 
fixed-time diffusion $t=0.5$ & 81.6 \% \\ 
no gradient features & 84.1 \% \\
unlearned gradient features & 85.6 \% \\
(full method) &  {\bftab 90.6 \%} \\
\bottomrule
\end{tabular}
\end{table}

To validate the components of our approach, we consider a simple ablation study on the full-resolution human segmentation task from \secref{segmentation}, using rotation-augmented raw coordinates as input. 
The variant \emph{no diffusion} omits the diffusion layer from each DiffusionNet block, \emph{fixed-time diffusion} manually specifies a diffusion time, \emph{no gradient features} omits the gradient features, and \emph{unlearned gradient features} includes gradient features but omits the learned transformation of gradient vectors $A$.
We observe a noticeable drop in accuracy when omitting any of the components of the method (\tabref{ablation}).
Manually specifying shared, non-optimal diffusion times ($t=0.1$, $t=0.5$) yields a network with significantly worse accuracy compared to our learned approach. 
A key advantage of our learned diffusion is that this time is automatically tuned by the optimization process, individually for each feature channel.

\paragraph{Spectral basis size}

\begin{figure}
\begin{center}
\end{center}
    \includegraphics{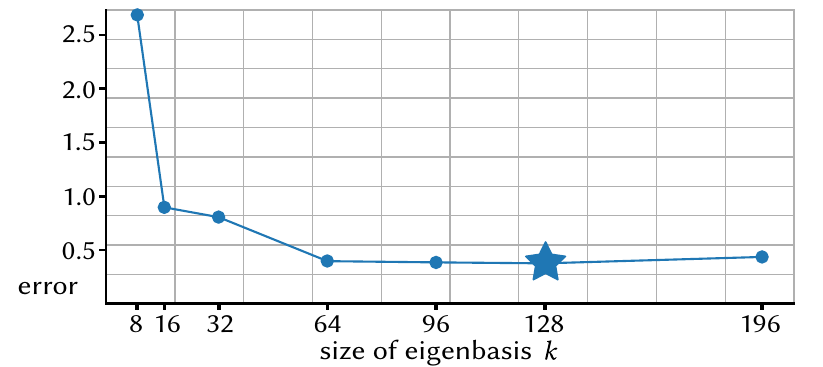}
    \caption{
      The effect of varying the size of the truncated basis for spectral diffusion evaluation, measured via error on the FAUST vertex-labelling correspondence from \tabref{sampling_invariance}, \figloc{orig}.
      We use $128$ eigenvectors in all experiments.
      \label{fig:eig_sweep_plot}
    }
\end{figure}

When evaluating diffusion with spectral acceleration (\secref{SpectralAcceleration}), increasing the size $k$ of the spectral basis more accurately resolves diffusion at the cost of increased computation.
In \figref{eig_sweep_plot} we vary $k$ for the FAUST vertex-labeling correspondence task as in \tabref{sampling_invariance}, measuring accuracy on the original test set.
We find performance degrades significantly with fewer than $64$ eigenvectors on this problem, while larger bases offer negligible benefit---our experiments use $k=128$ eigenvectors as a safe default.

\section{Experiment Details}
\label{app:details}

Here we provide additional methodology details for experiments.

\paragraph{Orientation}

\figref{leftright_segmentation} shows the results of a simple artificial experiment in which we segment the left vs. right side of human models from the FAUST dataset~\cite{bogo2014faust} using a purely intrinsic 32-width DiffusionNet with HKS as input.
On the original dataset, asymmetric biases--such as a template mesh with asymmetric connectivity--make it unintentionally easy to distinguish left from right.
We cancel the effect of these biases by augmenting the dataset with a copy of each mesh that has been mirrored across the left-right axis (preserving orientation by inverting triangles).
With a complex-valued $A$, our network is able to easily distinguish left from right with 99.9\% accuracy, despite both a purely intrinsic architecture and intrinsic input features.
Restricting to real-valued $A$ removes the effect; the network is unable to disambiguate the symmetry, with a totally random 50.0\% accuracy.

\paragraph{Human segmentation}

All results are given in \tabref{human-segmentation}.
Past work has used different variations of this dataset, both in terms of the input data and evaluation criteria. 
The original dataset presented by \citet{maron2017convolutional} contains moderately large meshes of up to 12k vertices, with segmentations labeled per-face, and accuracy is reported as the fraction of faces in the entire test set which were classified correctly.
The experiments from \citet{wiersma2020cnns} deviate slightly: they remap the ground truth to vertices, and train and test on a subsampling of the vertex set; nonetheless we group these results with the original dataset for the sake of simplicity as they are very similar.

MeshCNN \cite{hanocka2019meshcnn} generated a simplified version of the dataset where the meshes have $<1$k vertices, and segmentations have been remapped to edges.
Additionally, when reporting test evaluation, a soft ground truth is used allowing for multiple correct segmentation results for edges at the boundary between two regions.
For comparison we also apply DiffusionNet to this variant of the task, denoted by $^\dagger$ in \tabref{human-segmentation}---we directly generate a prediction per-edge by averaging per-vertex outputs to edges before applying the final softmax, and evaluate test results against the same soft ground truth.
Finally, PD-MeshNet \cite{milano2020primaldual} generated per-face labels for the MeshCNN simplified models and trained and tested on these \emph{without} any soft ground-truth---we denote this variant by $^\ddagger$ and again evaluate DiffusionNet with per-face predictions.

Across all variants, DiffusionNet achieves highly accurate performance.
Unlike many of these methods, DiffusionNet can easily be trained directly on the original meshes without any special treatment.
Even methods which evaluate on full-resolution models may be scalable only due to special pre- and post-processing schemes, which add complexity to adoption in practice---for instance, MeshWalker \cite{lahav2020meshwalker} trains on simplified meshes then applies an upsampling and smoothing scheme to handle full resolution data.

\paragraph{Discretization agnostic learning}

To investigate robustness to discretization on our remeshed FAUST dataset, we train several recent mesh-based and point-based surface learning methods, in addition to our own 256-width DiffusionNet with dropout.
For mesh-based methods, we also train SplineCNN~\cite{fey2018splinecnn}, ACSCNN~\cite{li2020shape}, HSN~\cite{wiersma2020cnns}; we also tried MeshCNN~\cite{hanocka2019meshcnn}, but found it prohibitively expensive.
For point-based methods, we train PointNet~\cite{qi2017pointnet} and DGCNN~\cite{wang2019dgcnn}, and consider using both the vertex set as a point set, as well as sampling a random point cloud on the surface, predicting there, then projecting the results back to vertices according to nearest-neighbors.
For equivalent comparison, all models are trained with only vertex positions as input (or the constant function, for ACSCNN and SplineCNN), and we augment during training with random rotations about the vertical axis to encourage rotation-invariance.
Wherever possible, we mimic the training configuration of the original work or make a best-effort to find suitable parameters for this task.
We note that some models perform slightly worse than previously reported results, presumably due to the use of simpler input features or learning in a rotation-invariant setting rather than aligned.

In general, only DiffusionNet learns accurate correspondences which are robust to remeshing and resampling.
In particular, ACSCNN still produces nearly perfect results on the original template meshes even in the rotation-invariant setting, but yields essentially random noise after any remeshing.
Perhaps unsurprisingly, point-based methods are less prone to overfitting the mesh connectivity (though the DGCNN on vertices still manages to do so), but are still notably less accurate than mesh-based techniques.
\figref{eval_correspondence} gives full geodesic error plots corresponding to \tabref{sampling_invariance}.

\end{document}